\documentclass{article}

\PassOptionsToPackage{numbers}{natbib}

    \usepackage[preprint]{neurips_2025}

\usepackage[utf8]{inputenc} 
\usepackage[T1]{fontenc}    
\usepackage{hyperref}       
\usepackage{url}            
\usepackage{booktabs}       
\usepackage{amsfonts}       
\usepackage{nicefrac}       
\usepackage{microtype}      
\usepackage{xcolor}         

\usepackage{multirow}
\usepackage{pifont}
\usepackage{bbding}
\usepackage{array}
\usepackage{tikz}
\usepackage{colortbl}   
\usepackage{epigraph}
\usepackage{wrapfig}
\usepackage{enumitem}
\usepackage{fontawesome5} 

\definecolor{gold}{RGB}{255, 215, 0} 
\definecolor{silver}{RGB}{192, 192, 192} 
\definecolor{bronze}{RGB}{205, 127, 50} 

\newcommand{\goldmedal}{{\color{gold}\faMedal}}
\newcommand{\silvermedal}{{\color{silver}\faMedal}}
\newcommand{\bronzemedal}{{\color{bronze}\faMedal}}

\title{X-MAS: Towards Building\\Multi-Agent Systems with Heterogeneous LLMs}

\author{%
  \textbf{Rui Ye\textsuperscript{1,*}} \quad \textbf{Xiangrui Liu\textsuperscript{1,*}} \\
  \textbf{Qimin Wu\textsuperscript{1}} \quad \textbf{Xianghe Pang\textsuperscript{1}} \quad \textbf{Zhenfei Yin\textsuperscript{2,3}}  \quad \textbf{Lei Bai\textsuperscript{4}} \quad \textbf{Siheng Chen\textsuperscript{1,†}} \\\\
  \textsuperscript{1} Shanghai Jiao Tong University \quad \textsuperscript{2} University of Oxford \quad \textsuperscript{3} The University of Sydney \\
  \textsuperscript{4} Shanghai AI Laboratory \quad \textsuperscript{*} Equal Contributions \quad \textsuperscript{†} Corresponding Author \\
  \textcolor{black}{X-MAS: \href{https://github.com/MASWorks/X-MAS}{https://github.com/MASWorks/X-MAS}}
}

\begin{document}

\maketitle

\begin{abstract}
  LLM-based multi-agent systems (MAS) extend the capabilities of single LLMs by enabling cooperation among multiple specialized agents.
  However, most existing MAS frameworks rely on a single LLM to drive all agents, constraining the system's intelligence to the limit of that model.
  This paper explores the paradigm of heterogeneous LLM-driven MAS (X-MAS), where agents are powered by diverse LLMs, elevating the system's potential to the collective intelligence of diverse LLMs.
  We introduce X-MAS-Bench, a comprehensive testbed designed to evaluate the performance of various LLMs across different domains and MAS-related functions.
  As an extensive empirical study, we assess 27 LLMs across 5 domains (encompassing 21 test sets) and 5 functions, conducting over 1.7 million evaluations to identify optimal model selections for each domain-function combination.
  Building on these findings, we demonstrate that transitioning from homogeneous to heterogeneous LLM-driven MAS can significantly enhance system performance without requiring structural redesign.
  Specifically, in a chatbot-only MAS scenario, the heterogeneous configuration yields up to 8.4\% performance improvement on the MATH dataset.
  In a mixed chatbot-reasoner scenario, the heterogeneous MAS could achieve a remarkable 47\% performance boost on the AIME dataset.
  Our results underscore the transformative potential of heterogeneous LLMs in MAS, highlighting a promising avenue for advancing scalable, collaborative AI systems.
\end{abstract}

\section{Introduction}

Large language models (LLMs) such as GPT~\cite{openai2023gpt4}, Gemini~\cite{team2024gemini}, Qwen~\cite{yang2024qwen2}, have been applied across various domains.
However, despite their remarkable capabilities, LLMs often struggle with multifaceted, complex, and real-world problems due to inherent limitations such as hallucinations~\cite{zhang2023siren,min2023factscore}.

In response to these limitations, LLM-based multi-agent systems (MAS) have emerged as a promising solution~\cite{masgpt,qian2024chatdev,co-scientist}.
MAS involves the collaboration of multiple agents, each specialized in specific functions, to address problems more effectively than a single model could.
his paradigm has been successfully applied across various scenarios, including software development~\cite{qian2024chatdev,hongmetagpt}, mathematics~\cite{leimacm,dylan}, and scientific discovery~\cite{boiko2023autonomous,aiscientists}.
For instance, ChatDev~\cite{qian2024chatdev}, MetaGPT~\cite{hongmetagpt}, and EvoMAC~\cite{hu2025selfevolving} utilize multiple coding agents (e.g., coders and testers) to improve software programming, while AI co-scientist~\cite{co-scientist} employs a MAS to enhance biomedical and scientific research.

Despite notable progress, most existing MAS frameworks rely on a single LLM to drive all agents~\cite{hongmetagpt,qian2024chatdev,dylan,masgpt,hu2025selfevolving,debate1,chen2023agentverse}.
This manner inherently limits the system’s intelligence to that of the underlying model.
For example, if a single LLM produces fundamental errors in certain facts, these mistakes are unlikely to be corrected through the collaboration of agents powered by the same model.
Inspired by the advantages of diversity in collective intelligence~\cite{hong2004groups,kozhevnikov2014cognitive,aggarwal2015cognitive}, this paper explores MAS with heterogeneous LLMs (X-MAS), pushing the system’s capabilities beyond its previous limit to harness the collective potential of LLMs trained on diverse corpora or by different teams~\cite{qwen2.5,dubey2024llama,qwen-math}.

To provide a comprehensive evaluation of LLMs in MAS, we introduce \textbf{X-MAS-Bench}, a testbed designed to assess the performance of various LLMs across different MAS-related functions and domains. 
Specifically, we consider 5 representative functions of agents in MAS, including question-answering~\cite{debate1,hongmetagpt}, revise~\cite{hu2025selfevolving,Self-refine}, aggregation~\cite{macnet,debate2}, planning~\cite{islam2024mapcoder,leimacm}, and evaluation~\cite{chen2023agentverse,qian2024chatdev}; as well as 5 common domains, including mathematics, coding, science, medicine, and finance—spanning 21 test sets.
Each function is assessed under controlled experimental conditions.
For example, when assessing aggregation, each query is input into several pre-defined LLMs, whose outputs are concatenated to be aggregated by the examined LLM.
The aggregated responses of various LLMs are then evaluated and compared.
Finally, we assess 27 LLMs across these 5 functions and 5 domains, conducting over 1.7 million evaluations to identify the optimal model selections for each domain-function combination.
Our findings include that
(1) no single LLM excels across all scenarios,
(2) a single LLM could have significant performance variation across functions and domains,
(3) different LLMs may show large performance disparities within the same function and domain,
(4) smaller LLMs can sometimes outperform larger ones,
highlighting the potential advantages of employing heterogeneous LLMs in MAS.
These results provide valuable insights for researchers and practitioners in selecting the most appropriate LLMs for their specific MAS applications.

\begin{figure}[t]
    \centering
    \includegraphics[width=1.0\linewidth]{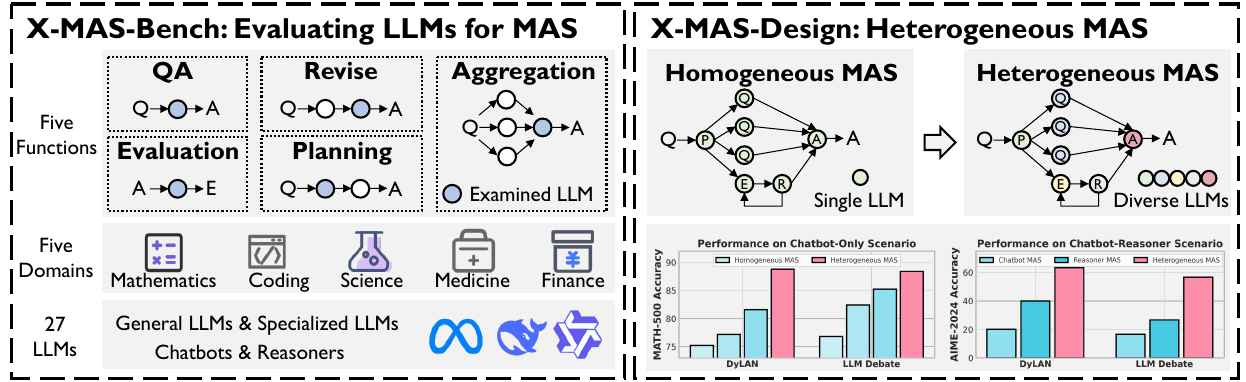}
    \vspace{-4mm}
    \caption{Overview of our X-MAS-Bench and X-MAS-Design. X-MAS-Bench assesses the capabilities of LLMs in MAS while X-MAS-Design focuses on transitioning a homogeneous MAS to a heterogeneous one, gaining from the observations in X-MAS-Bench. Experiments on chatbot-only and mixed chatbot-reasoner scenarios evidently show the benefits of heterogeneous MAS.}
    \vspace{-3mm}
    \label{fig:overview}
\end{figure}

Building on these observations, we explore the effects of transitioning from homogeneous to heterogeneous LLM-driven MAS (\textbf{X-MAS-Design}).
As a proof of concept, given the implementation of a MAS method, we simply assign agents with appropriate LLMs (taking seconds) by referring observations in X-MAS-Bench.
To validate our idea, we examine three existing MAS frameworks—LLM-Debate~\cite{debate1}, AgentVerse~\cite{chen2023agentverse}, and DyLAN~\cite{dylan}—as well as a prototype MAS designed by us, which incorporates all five functions in one system.
Our analysis covers five domains with no sample overlap compared to X-MAS-Bench.
In a chatbot-only scenario, we observe consistent improvements in performance for heterogeneous MAS over homogeneous configurations, achieving up to a 8.4\% performance gain on the MATH~\cite{hendrycksmath2021} benchmark.
Interestingly, while reasoner-only MAS often underperforms relative to chatbot-only systems, combining chatbot and reasoner in a heterogeneous MAS leads to significant performance improvements.
Specifically, in the competition-level AIME-2024 benchmark, AgentVerse~\cite{chen2023agentverse} is improved from 20\% to 50\%, and DyLAN~\cite{dylan} improved from 40\% to 63\%.
Our further experiments reveal that increasing the number of candidate LLMs for heterogeneous MAS results in a monotonic performance improvement, reinforcing the value of LLM diversity in MAS.
Based on our work, future research could explore more nuanced strategies for selecting and integrating LLMs in heterogeneous MAS;
investigate the scalability and adaptability of heterogeneous MAS across different industries and other complex tasks.

Our contributions are as follows:
\begin{enumerate}[leftmargin=*]
    \item \textbf{X-MAS-Bench:} We assess 27 LLMs across 5 MAS-related functions and 5 domains, conducting over 1.7 million evaluations to identify diverse optimal model selections for each domain-function combination. These observations could benefit researchers and practitioners in building MAS.
    \item \textbf{X-MAS-Design:} Based on findings in X-MAS-Bench, we propose to transition existing MAS methods from homogeneous to heterogeneous LLM-driven MAS. We conduct extensive experiments, showing that heterogeneous MAS consistently outperforms homogeneous MAS.
    \item \textbf{Open Source:} We release all data, code, and evaluation results to facilitate future MAS research.
\end{enumerate}

\section{Related Work}

\textbf{LLM-based MAS.}
LLM-based multi-agent systems (MAS) leverage multiple LLM-based agents to collaborate for better task solving than single LLM~\cite{chen2023agentverse,hongmetagpt,adas,masgpt}.
ChatDev~\cite{qian2024chatdev}, MetaGPT~\cite{hongmetagpt}, and EvoMAC~\cite{hu2025selfevolving} use multiple coding agents (e.g., coders and testers) for software programming;
while MACM~\cite{leimacm} applies math agents for mathematics.
Focusing on general tasks, debate-based methods~\cite{debate1,debate2} enable multiple experts in debating for better solutions;
AgentVerse~\cite{chen2023agentverse} and DyLAN~\cite{dylan} dynamically adjust the agent team for task solving;
while MAS-GPT~\cite{masgpt} trains an LLM for generating MAS.
However, all of these methods rely on a single LLM to drive all agents, which inherently limits the system's intelligence to that of the underlying LLM.
This paper proposes to push the limit by harnessing the collective intelligence of heterogeneous LLMs from different sources.

\textbf{Heterogeneous LLMs.}
In a general context of LLMs, there are several works related to using heterogeneous LLMs~\cite{chen2023less,venkatraman2024collabstory}.
LLM-Blender~\cite{llm-blender} trains a model for ensembling outputs from multiple LLMs.
MoA~\cite{moa} and ReConcile~\cite{chen2024reconcile} enable multiple LLMs for discussion, however, involving all candidate LLMs without considering their appropriateness.
MASRouter~\cite{yue2025masrouter} manually selects several candidate LLMs for MAS and is optimized for their specific framework.
In contrast, our paper systematically assess the capabilities of LLMs across several MAS-related functions and domains, aiming to universally benefit the design of heterogeneous MAS for various MAS methods.

\textbf{Benchmarking LLMs.}
Many works benchmark the capabilities of LLMs in various domains (such as math~\cite{hendrycksmath2021}, coding~\cite{jimenezswe}, science~\cite{rein2023gpqa}, medicine~\cite{healthbench}, and finance~\cite{xie2023pixiu}) and functions (such as planning~\cite{valmeekam2023planbench} and evaluation~\cite{tan2025judgebench}).
However, our paper for the first time benchmarks LLMs for MAS, which assesses the capabilities of LLMs across 25 function-domain perspectives related to MAS.

\section{X-MAS-Bench: Evaluating LLMs for MAS}

X-MAS-Bench is a testbed designed to assess the performance of various LLMs across different MAS-related functions and domains.
Specifically, we consider 5 representative functions of agents in MAS—question-answering~\cite{debate1,hongmetagpt}, revise~\cite{hu2025selfevolving,Self-refine}, aggregation~\cite{macnet,debate2}, planning~\cite{islam2024mapcoder,leimacm}, and evaluation~\cite{chen2023agentverse,qian2024chatdev}.
Orthogonally, we investigate behaviors in 5 domains, including mathematics, coding, science, medicine, and finance—spanning 21 test sets.
Each function is assessed under controlled experimental conditions.
In this section, we demonstrate the details of experimental conditions in Section~\ref{sec:details_functions} and experimental results in Section~\ref{sec:bench_experiments}.

\subsection{Benchmarking MAS-Related Functions}
\label{sec:details_functions}

To systematically assess LLM capabilities in a multi-agent context, we decompose MAS behaviors into five representative and commonly-used agent functions: question-answering, revise, aggregation, planning, and evaluation. For each function, we define a standardized prompt protocol. In all cases, we carefully control experimental conditions such that the only varying factor is the LLM under evaluation. Below we detail the designed assessment of each function.

\textbf{Question-answering.} The question‑answering (QA) function measures a LLM’s ability to comprehend a question and produce a correct answer in free‑text form.
This function is fundamental to all MAS methods such as LLM‑Debate~\cite{debate1}, where each agent first independently answers the query, and to MetaGPT~\cite{hongmetagpt}, in which an agent supplies initial draft responses in first stage of the processing.
In QA assessment, LLM receives a sampled query from any test dataset (e.g., MATH~\cite{hendrycksmath2021}) as input and returns an answer in free‐text format.
The outputs are compared with ground-truth answers, where the resulted accuracy will be the indicator to assess the QA ability.

\textbf{Revise.}
The function of revise assesses an agent’s capacity to revise an initial answer, which is potentially flawed, to produce a corrected answer.
This function commonly exists in workflows requiring iterative refinement, such as the update agent in EvoMAC~\cite{hu2025selfevolving} and the refiner in Self-Refine~\cite{Self-refine}.
In assessment of revise, given a prompt formatted with a sampled query and a corresponding answer generated by a pre-defined LLM, the examined LLM is asked to provide a final complete answer by reasoning and revising over the provided query and answer.
In this case, all examined LLMs are provided with the same prompt (i.e., sourced from the same query and same pre-defined LLM), therefore ensuring fair comparisons among examined LLMs.
Similarly, the revised outputs are compared with ground-truth answers and the resulting accuracy denotes the revise capability.

\textbf{Aggregation.}
Aggregation refers to the capability of combining multiple candidate answers into a coherent, correct, and even improved final answer.
It is a key mechanism in MAS that leverages multiple answering paths, such as convergent agents in MacNet~\cite{macnet}, and extractive mode of the judge in MAD~\cite{debate2}.
To assess aggregation, for each query, we collect candidate responses from a fixed set of pre-defined LLMs (set to 3 in this case).
The examined LLM is provided with the query and these candidate responses in a fixed concatenated format, and is asked to synthesize the final answer.
Importantly, all candidate responses and their order remain identical across all models, ensuring consistency in the prompts and allowing for fair comparisons.
The aggregated answers are then scored against the ground-truth answers using accuracy as the metric.

\textbf{Planning.}
Planning involves decomposing a task into sub-tasks and assigning appropriate roles to agents along a workflow to solve the problem collaboratively.
This function is critical in systems like MACM~\cite{leimacm} and MapCoder~\cite{islam2024mapcoder}, where a thinker or a planning agent defines the entire agentic workflow.
In planning assessment, the task of the examined LLM is to provide a suitable plan to the sampled query by giving the role descriptions of the agents required to answer the query together with the workflow, whose output should follow a pre-defined format for subsequent string extraction.
Subsequently, the ordered role descriptions and the number of roles are extracted.
Next, the corresponding number (i.e., role number) of candidate LLMs are activated for action according to the role description and workflow arrangement.
Input prompts and candidate LLMs are kept fixed across all examined LLMs.
The overall system performance—evaluated by final task accuracy—serves as the proxy for planning capability.

\begin{figure}[t]
    \centering
    \includegraphics[width=1.0\linewidth]{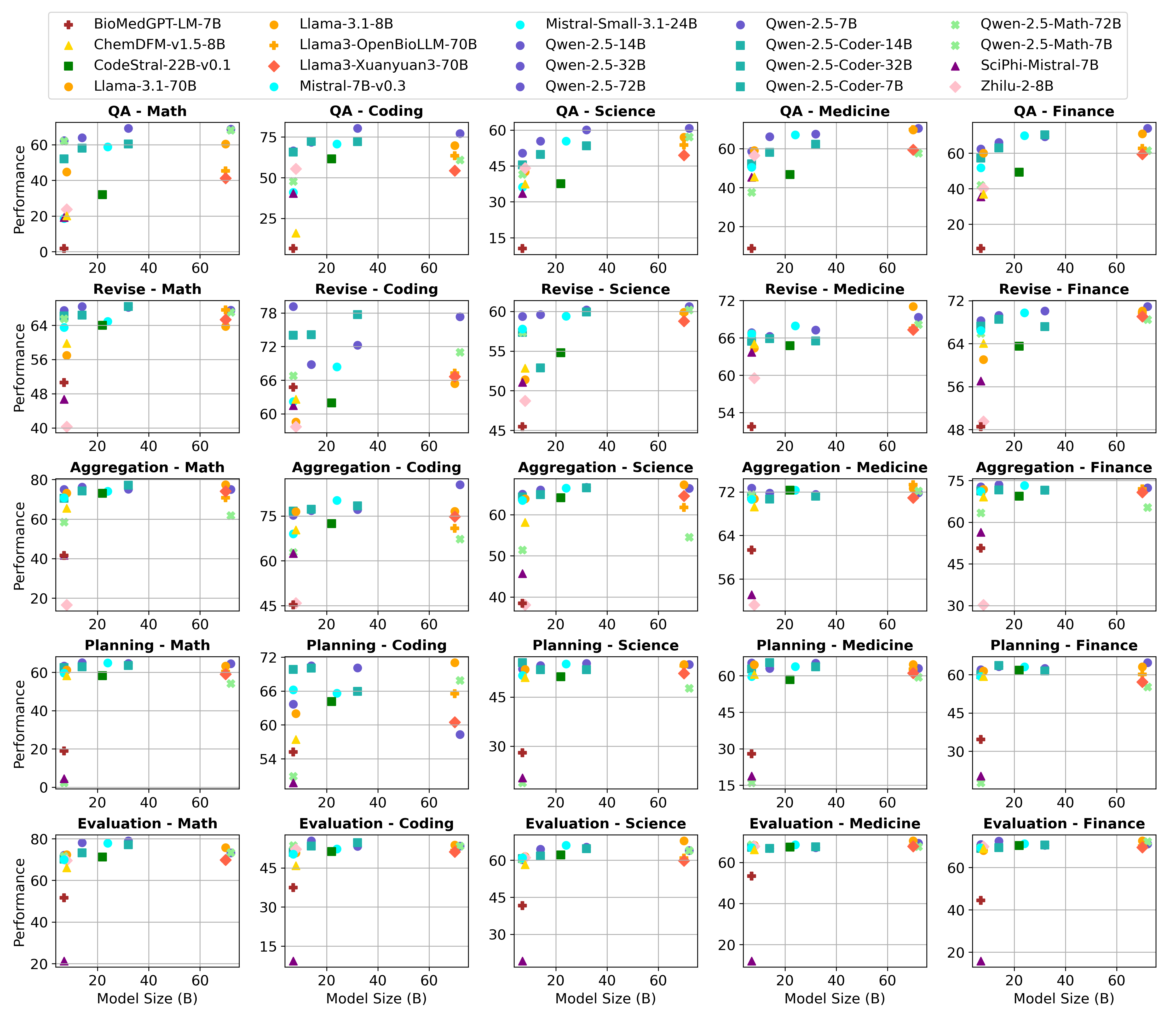}
    \vspace{-3mm}
    \caption{Benchmarking chatbot LLMs on 5 MAS-related functions and 5 domains. We see that no single LLM excels across all scenarios, indicating the potential advantages of employing heterogeneous LLMs in MAS. All evaluation results will be open-sourced for future research.}
    \vspace{-3mm}
    \label{fig:bench_results}
\end{figure}

\textbf{Evaluation.}
Evaluation measures an agent’s ability to critically assess the quality or correctness of other agents’ outputs.
This function is used in MAS to filter faulty reasoning, select better solutions, determine early stopping, or guide further actions—commonly seen in frameworks like AgentVerse~\cite{chen2023agentverse} and ChatDev~\cite{qian2024chatdev}.
In our assessment, each examined LLM is presented with a query and an answer generated by a pre-defined LLM.
The examined LLM is instructed to determine whether the provided answer correctly address the query.
The candidate answer and evaluation instruction remain constant across LLMs to ensure fair comparison.
Unlike the evaluation in previous functions, the judgment of the examined LLM is compared against ground-truth correctness.

\subsection{Experiments in Evaluating LLMs across Functions and Domains}
\label{sec:bench_experiments}

Following the above definitions of functions, this section assesses the capabilities of various LLMs in different functions and domains, aiming at demonstrating the landscape of LLMs for MAS.
The reported results are expected to demonstrate the potential of leveraging heterogeneous LLMs for MAS and facilitate future researchers in choosing appropriate LLMs for their MAS.

\textbf{Experimental setups.} 
We examine 27 LLMs, covering 20 chatbots (i.e., instructed LLMs) and 7 reasoners (i.e., reasoning LLMs).
Among the 20 chatbots, we consider general chatbots trained by different companies such as Llama~\cite{dubey2024llama}, Qwen~\cite{qwen2.5}, Mistral~\cite{Mistral-7B-Instruct-v0.3,Mistral-Small-3.1-24B-Instruct-2503}, and domain-specific chatbots including mathematics~\cite{qwen-math}, coding~\cite{Qwen2.5-coder}, science~\cite{chemdfm,SciPhi-Mistral-7B-32k}, medicine~\cite{Llama3-XuanYuan3-70B-Chat,ZhiLu-2-8B-Instruct}, and finance~\cite{Llama3-XuanYuan3-70B-Chat,ZhiLu-2-8B-Instruct}.
The reasoners include LLMs from DeepSeek~\cite{guo2025deepseek}, Qwen~\cite{qwq32b}, and others~\cite{openthoughts}.
We set each model’s maximum token limit to its own capacity, 8192 tokens maxed, with a temperature of 0.5 by default.
Specially, all LLMs instantiated within the planning workflow are executed with their temperature fixed at 0 to guarantee as the planning involves format-following.
Our datasets cover domains including mathematics~\cite{hendrycksmath2021,aqua-rat,gsm-hard,AIME-2024,mmlu,wang2024mmlupro}, coding~\cite{humaneval,mbpp,evalplus}, science~\cite{rein2023gpqa,wangscibench,sun2024scieval,feng2024sciknoweval}, medicine~\cite{pal2022medmcqa,medqa,jin2019pubmedqa}, and finance~\cite{islam2023financebench,chen2021finqa,fpb}, where each dataset is randomly sampled up to 500 examples without replacement; see more details in Section~\ref{app:experimental_setups}.

\begin{table}[t]
  \centering
  \caption{Summary of top-3 LLMs for each function-domain combination (chatbot-only scenario). All the assessed LLMs are instructed models (e.g., Qwen2.5-32B denotes Qwen2.5-32B-Instruct.). We see that no single LLM excels across all scenarios. Meanwhile, the top models are not always those with the largest sizes, indicating the potential of improving both performance and cost.}
  \label{tab:model-ranking}
  \renewcommand{\arraystretch}{1.2}
  \resizebox{\textwidth}{!}{
  \begin{tabular}{l c c c c c c}
    \toprule
    \textbf{Function} & \textbf{Rank}
     & \textbf{Mathematics} & \textbf{Coding} & \textbf{Science}  
     & \textbf{Medicine}  & \textbf{Finance} \\
    \midrule

  \rowcolor{gray!10}
    & \goldmedal  & Qwen2.5-32B (69.2)
                 & Qwen2.5-32B (80.3)
                 & Qwen2.5-72B (60.7)
                 & Qwen2.5-72B (70.4)
                 & Qwen2.5-72B (74.0) \\
  \rowcolor{gray!10}  
    \textbf{QA}
    & \silvermedal& Qwen2.5-72B (68.8)
                 & Qwen2.5-72B (77.1)
                 & Qwen2.5-32B (60.0)
                 & Llama3-OpenBioLLM-70B (69.7)
                 & Qwen2.5-32B (71.0) \\
  \rowcolor{gray!10}
    & \bronzemedal & Qwen2.5-Math-72B (68.2)
                  & Qwen2.5-Coder-14B (72.3)
                  & Qwen2.5-Math-72B (57.1)
                  & Llama-3.1-70B (69.6)
                  & Qwen2.5-Coder-32B (70.3) \\
    \midrule

    & \goldmedal  & Qwen2.5-Coder-32B (68.4)
                 & Qwen2.5-7B (79.2)
                 & Qwen2.5-72B (60.6) 
                 & Llama-3.1-70B (71.0)
                 & Qwen2.5-72B (70.9) \\
    \textbf{Revise}
    & \silvermedal& Qwen2.5-14B (68.4)
                 & Qwen2.5-Coder-32B (77.7)
                 & Qwen2.5-32B (60.2)
                 & Qwen2.5-72B (69.3)
                 & Llama-3.1-70B (70.1) \\
    & \bronzemedal & Qwen2.5-32B (68.2)
                  & Qwen2.5-72B (77.3)
                  & Qwen2.5-Math-72B (60.2)
                  & Qwen2.5-Math-72B (68.1)
                  & Qwen2.5-32B (70.1) \\
    \midrule

  \rowcolor{gray!10}
    & \goldmedal  & Llama-3.1-70B (77.4)
                 & Qwen2.5-72B (85.5)
                 & Llama-3.1-70B (67.3)
                 & Llama3-OpenBioLLM-70B (73.4)
                 & Qwen2.5-14B (73.6) \\
  \rowcolor{gray!10}
    \textbf{Aggregation}
    & \silvermedal& Qwen2.5-Coder-32B (77.1)
                 & Mistral-Small-3.1-24B (80.2)
                 & Qwen2.5-32B (66.7)
                 & Qwen2.5-7B (72.7)
                 & Mistral-Small-3.1-24B (73.2) \\
  \rowcolor{gray!10}
    & \bronzemedal & Qwen2.5-14B (76.2)
                  & Qwen2.5-Coder-32B (78.4)
                  & Qwen2.5-Coder-32B (66.5)
                  & Llama-3.1-70B (72.7)
                  & Qwen2.5-7B (72.8) \\
    \midrule

    & \goldmedal  & Qwen2.5-14B (65.0)
                 & Llama-3.1-70B (71.0)
                 & Qwen2.5-Coder-7B (55.5)
                 & Qwen2.5-Coder-14B (65.4)
                 & Qwen2.5-72B (64.7) \\
    \textbf{Planning}
    & \silvermedal& Mistral-Small-3.1-24B (65.0)
                 & Qwen2.5-14B (70.5)
                 & Qwen2.5-32B (55.3)
                 & Qwen2.5-7B (65.3)
                 & Qwen2.5-Coder-14B (63.6) \\
    & \bronzemedal & Qwen2.5-32B (64.7)
                  & Qwen2.5-32B (70.1)
                  & Mistral-Small-3.1-24B (55.1)
                  & Qwen2.5-32B (65.2)
                  & Qwen2.5-14B (63.2) \\
    \midrule
    
  \rowcolor{gray!10}
    & \goldmedal  & Qwen2.5-32B (79.0)
                 & Qwen2.5-14B (55.4)
                 & Llama-3.1-70B (67.9)
                 & Llama-3.1-70B (70.5)
                 & Llama-3.1-70B (72.6) \\
  \rowcolor{gray!10}
    \textbf{Evaluation}
    & \silvermedal& Qwen2.5-14B (78.1)
                 & Qwen2.5-Coder-32B (54.7)
                 & Mistral-Small-3.1-24B (66.1)
                 & Qwen2.5-72B (69.4)
                 & Qwen2.5-14B (72.6) \\
  \rowcolor{gray!10}
    & \bronzemedal & Mistral-Small-3.1-24B (77.9)
                  & Llama-3.1-70B (53.8)
                  & Qwen2.5-32B (65.3)
                  & Mistral-Small-3.1-24B (68.7)
                  & Qwen2.5-Math-72B (72.3) \\
    \bottomrule
  \end{tabular}
\vspace{-3mm}
  }
\end{table}

\textbf{No single LLM excels across all scenarios.}
We plot the size-performance values of each evaluated chatbot LLM across 25 function-domain combinations in Figure~\ref{fig:bench_results} and report the summary of top-3 LLMs for each combination in Table~\ref{tab:model-ranking}; see results of all LLMs in Figure~\ref{fig:bench_results_full} and Table~\ref{tab:model-ranking_full}.
From these results, we see that
(1) No single LLM excels universally across all scenarios.
A heterogeneous MAS can capitalize on these differences by assigning scenario-specialized models (e.g., Llama3-OpenBioLLM for medicine) to specific agents, maximizing collective intelligence.
(2) LLMs exhibit varied performance across MAS-related functions, reinforcing the value of heterogeneity.

\textbf{A single LLM could have significant performance variation across domains and functions.}
Individual LLMs exhibit substantial performance disparities when evaluated across different domains and functions, underscoring the limitations of relying on a single model in a homogeneous MAS.
For instance, in Figure~\ref{fig:bench_results}, Qwen2.5-7B performs exceptionally well for revising in coding domain; while dropping to a mid-tier level for revising in medicine domain and planning in coding domain.

\textbf{There are large performance disparities across LLMs within the same domain and function.}
For the function of revise or the domain of coding, we observe diverse behaviors on the examined LLMs, as shown by disperse scatters in Figure~\ref{fig:bench_results} (second row and second column).

\textbf{Smaller LLMs can outperform larger ones in niche scenarios.}
While larger models like Qwen2.5-72B-Instruct and Llama-3.1-70B-Instruct often lead, smaller models occasionally excel in specific function-domain pairs.
For example, in revise-coding pair, Qwen2.5-7B-Instruct (79.2) outperforms Qwen2.5-72B-Instruct (77.3); while in aggregation-finance and evaluation-finance pairs, Qwen2.5-14B achieves the best performance among all models.
This indicates that heterogeneous MAS can incorporate smaller, specialized models to optimize performance and computational efficiency, reducing reliance on resource-intensive large models while maintaining or improving outcomes.

\textbf{Low-performing models highlight the risk of homogeneous MAS.}
Some models consistently underperform across domains and functions (e.g., BioMedGPT-LM-7B and SciPhi-Mistral-7B-32k).
A homogeneous MAS relying on such models would be severely limited, whereas a heterogeneous setup can mitigate this by integrating appropriate and high-performing LLMs.

\textbf{Consistent high performers enable robust heterogeneous configurations.}
Models like Qwen-2.5-32B-Instruct, Qwen-2.5-72B-Instruct, and Llama-3.1-70B-Instruct frequently rank among the top across domains and functions (e.g., 80.3 in QA-coding, 79.0 in evaluation-math for Qwen-2.5-32B-Instruct).
These models can serve as reliable anchors in a heterogeneous MAS, complemented by specialized models for niche tasks (e.g., Llama3-OpenBioLLM-70B in medicine), ensuring robust and scalable performance improvements.

\section{X-MAS-Design: Leveraging Diversity for MAS}

Based on the findings in X-MAS-Bench (Section~\ref{sec:bench_experiments}), we explore the effects of transitioning from homogeneous to heterogeneous LLM-driven MAS (X-MAS-Design).
We show how a homogeneous MAS is transformed into a heterogeneous MAS in Section~\ref{sec:transitioning_mas}.
We provide experimental results in a chatbot-only scenario (Section~\ref{sec:mas_experiments_chatbot}) and a mixed chatbot-reasoner scenario (Section~\ref{sec:mas_experiments_mixed}).

\subsection{Transitioning from Homogeneous to Heterogeneous LLM-driven MAS}
\label{sec:transitioning_mas}

\textbf{Transitioning existing MAS methods.}
As a proof of concept, we aim to show that a simple manual modification of the LLM configurations can enhance the performance of MAS without any structural improvement.
For each target MAS method (e.g., AgentVerse~\cite{chen2023agentverse}, LLM-Debate~\cite{debate1}), we retain the original agent roles and interaction topology but substitute the single homogeneous LLM with several appropriate LLMs for the agents.
Concretely, for each domain-function pair in the original design (e.g., the evaluator for coding in AgentVerse), we replace the uniform LLM driver with the top performer in the pool of available models based on observations from X-MAS-Bench (Section~\ref{sec:bench_experiments}).
By preserving the method's interaction logic and prompt templates, we ensure that any performance gains stem solely from LLM heterogeneity rather than modifications of workflow.
Please note that this modification is efficient as it only takes human researchers less than one minute to accomplish and could be automated even if we replace humans with LLMs with limited sizes (e.g., 7B~\cite{qwen2.5}).

\textbf{X-MAS-Proto.}
In addition to adapting existing MAS methods to heterogeneous ones, we implement X‑MAS‑Proto, a prototype MAS that explicitly implements all five functions (QA, revise, aggregation, planning, evaluation) in a single pipeline, serving as a proper object for investigation.
The system (see the MAS in Figure~\ref{fig:overview}) first invokes a planning agent to generate several different high-level ideas to the question;
next, multiple QA agents concurrently answer the query based on its corresponding ideas while one of the answers will be evaluated and revised to obtain a potentially better answer;
finally, an aggregation agent synthesizes across answers to get the final solution.
With X-MAS-Proto, we could straightforwardly assign appropriate LLMs for different functional agents, aiming to clearly demonstrate the benefits of LLM heterogeneity in MAS.

\begin{table}[t]
    \centering
    \caption{Transitioning from homogeneous to heterogeneous LLM-driven MAS (X-MAS-Design). There are four considered MAS methods and four candidate models. X-MAS-Design consistently achieves top performances across 5 domains (3 are relatively \colorbox{orange!15}{out-of-domain} for candidate LLMs).}
    \label{tab:x_mas_design_chatbot}
    \resizebox{\textwidth}{!}{
    \begin{tabular}{cc|cccccc}
    \toprule
        MAS Method & LLM & Math & Coding & \cellcolor{orange!15}Science & \cellcolor{orange!15}Medicine & \cellcolor{orange!15}Finance & Average \\
    \midrule
         & Qwen2.5-Math-7B & 2.40 & 3.21 & 0.40 & 6.00 & 5.33 & 3.47\\
         & Qwen2.5-Coder-32B & 75.20 & 72.69 & 32.00 & 47.60 & 64.00 & 58.30\\
         & Qwen2.5-32B & 83.20 & 76.31 & 34.00 & 50.40 & \textbf{74.67} & 63.72\\
         & Mistral-3.1-24B & 66.80 & 62.25 & 31.20 & 40.00 & 65.33 & 55.12\\
         \rowcolor{blue!8} \multirow{-5}{*}{\cellcolor{white}AgentVerse~\cite{chen2023agentverse}} & \textbf{X-MAS-Design} & \textbf{88.40} & \textbf{77.51} & \textbf{41.20} & \textbf{51.20} & 72.00 & \textbf{66.06}\\
    \midrule
        & Qwen2.5-Math-7B & 79.20 & 40.96 & 29.60 & 35.20 & 30.67 & 43.13\\
        & Qwen2.5-Coder-32B & 82.40 & 78.71 & 34.40 & 46.80 & 68.00 & 62.06\\
        & Qwen2.5-32B & 85.20 & 75.50 & 32.80 & 50.80 & 77.33 & 64.33\\
        & Mistral-3.1-24B & 76.80 & 66.67 & 33.60 & \textbf{52.00} & 66.67 & 59.15\\
        \rowcolor{blue!8} \multirow{-5}{*}{\cellcolor{white}LLM-Debate~\cite{debate1}} & \textbf{X-MAS-Design} & \textbf{88.40} & \textbf{79.92} & \textbf{39.20} & 51.60 & \textbf{77.33} & \textbf{67.29}\\
    \midrule
        & Qwen2.5-Math-7B & 0.00 & 13.25 & 15.20 & 13.20 & 5.33 & 9.40\\
        & Qwen2.5-Coder-32B & 77.20 & 78.31 & 34.80 & 41.60 & 61.33 & 58.65\\
        & Qwen2.5-32B & 81.60 & 74.70 & 38.00 & 46.00 & 73.33 & 62.73\\
        & Mistral-3.1-24B & 75.20 & 61.85 & 32.80 & 41.60 & 72.00 & 56.69\\
        \rowcolor{blue!8} \multirow{-5}{*}{\cellcolor{white}DyLAN~\cite{dylan}} & \textbf{X-MAS-Design} & \textbf{88.80} & \textbf{78.71} & \textbf{38.80} & \textbf{47.20} & \textbf{76.00} & \textbf{65.90}\\
    \midrule
        & Qwen2.5-Math-7B & 10.40 & 12.85 & 2.00 & 10.80 & 5.33 & 8.28\\
        & Qwen2.5-Coder-32B & 82.00 & 76.71 & 33.60 & 46.80 & 58.67 & 59.56\\
        & Qwen2.5-32B & 82.00 & 69.88 & 31.20 & 45.60 & 72.00 & 60.14\\
        & Mistral-3.1-24B & 78.80 & 63.05 & 34.40 & 46.40 & 72.00 & 58.93\\
        \rowcolor{blue!8} \multirow{-5}{*}{\cellcolor{white}X-MAS-Proto} & \textbf{X-MAS-Design} & \textbf{90.40} & \textbf{78.71} & \textbf{40.00} & \textbf{46.80} & \textbf{73.33} & \textbf{65.85}\\
    \bottomrule
    \end{tabular}
    \vspace{-5mm}
    }
\end{table}

\subsection{Experiments in Chatbot-only Scenarios}
\label{sec:mas_experiments_chatbot}

\textbf{Experimental setups.}
We experiment on X-MAS-Proto and three existing MAS methods including AgentVerse~\cite{chen2023agentverse}, LLM-Debate~\cite{debate1}, and DyLAN~\cite{dylan}.
Considering performances and efficiencies, we select four candidate chatbot LLMs: Qwen‑2.5‑32B, Mistral‑Small‑3.1‑24B, Qwen‑2.5‑Coder‑32B, and Qwen‑2.5‑Math‑7B.
We test MAS on a \textit{held-out} test splits of MATH‑500, MBPP, SciBench, PubMedQA, and FinanceBench, covering the examined 5 domains.
See model selection in Section~\ref{app:experiments_mas_chatbot}.


\textbf{Consistent performance gains of X-MAS-Design over homogeneous MAS.}
Table~\ref{tab:x_mas_design_chatbot} reports the performance comparisons of the homogeneous and heterogeneous versions of four MAS methods, where four LLMs are selected as candidates.
The table demonstrates that X-MAS-Design, the heterogeneous MAS configuration, consistently outperforms all homogeneous configurations on average for four methods.
In DyLAN, X-MAS-Design achieves an average performance of 65.90, surpassing the best homogeneous model (Qwen2.5-32B, 62.73) by 3 points.
There are only two outlier cases—LLM-Debate in medicine and Agentverse in finance—likely due to the candidate LLMs not including specialized models for these particular domains.
These results validate the X-MAS-Bench findings, which identified optimal model selections for domain-function combinations.
By leveraging diverse and appropriate LLMs, X-MAS-Design harnesses collective intelligence, leading to superior performance without requiring structural changes to existing MAS methods.

\textbf{Method-agnostic benefits of heterogeneity.}
The performance improvements of X-MAS-Design are consistent across all four MAS methods, despite their differing architectures and philosophies.
This method-agnostic nature of X-MAS-Design’s improvements highlights its versatility, providing strong evidence of our core idea in advocating X-MAS.

\textbf{X-MAS-Design could leverage the strengths of weak models to offset their weaknesses.}
Homogeneous configurations show significant variability in performance across domains, with certain models underperforming in specific areas.
For example, Qwen2.5-Math-7B performs poorly in most domains (e.g., 2.40 in Math, 0.40 in Science for AgentVerse), indicating its limited generalizability.
Even stronger models like Qwen2.5-32B and Mistral-3.1-24B show weaknesses, such as Mistral-3.1-24B’s 31.2 in Science (AgentVerse) or Qwen2.5-32B’s 31.2 in Science (X-MAS-Proto).
In contrast, X-MAS-Design consistently achieves balanced performance.
That is, X-MAS-Design mitigates the limitations of individual LLMs by combining their strengths, indicating the benefits of collective intelligence and that our X-MAS-Bench provides helpful guidance for the design of X-MAS.

\subsection{Experiments in Mixed Chatbot-Reasoner Scenarios}
\label{sec:mas_experiments_mixed}

\textbf{Experimental setups.}
The examined MAS methods follow that in Section~\ref{sec:mas_experiments_chatbot}.
As chatbots and reasoners exhibit different behaviors, we consider two candidate LLMs: Qwen-2.5-72B-Instruct and DeepSeek-R1-Distill-Qwen-32B.
These methods are tested on AIME-2024 and \textit{held-out} splits of MBPP, SciBench, PubMedQA and FinanceBench, covering the five examined domains.
We also test the methods on \textit{entirely new} (compared to X-MAS-Bench) test sets: AIME-2025~\cite{AIME2025} (the latest AIME math competition) and MATH-MAS~\cite{zhou2025reso} (multi-step).
See model selection in Section~\ref{app:experiments_mas_mixed}.

\begin{table}[t]
    \centering
    \caption{Effectiveness of X-MAS-Design in mixing chatbots and reasoners. While reasoner-based homogeneous MAS performs worse than chatbot-based homogeneous MAS, incorporating chatbots and reasoners into heterogeneous MAS contributing to large performance improvement.}
    \label{tab:x_mas_design_mixed}
    \resizebox{\textwidth}{!}{
    \begin{tabular}{cc|cccccc}
    \toprule
        MAS Method & LLM & Math & Coding & Science & Medicine & Finance & Average \\
    \midrule
        & Chatbot & 20.00 & 75.50 & 37.60 & 47.20 & 72.00 & 50.46\\
        & Reasoner & 0.00 & 11.65 & 5.60 & 44.40 & 21.33 & 16.60\\
        \rowcolor{blue!8} \multirow{-3}{*}{\cellcolor{white}AgentVerse~\cite{chen2023agentverse}} & \textbf{X-MAS-Design} & \textbf{50.00} & \textbf{77.91} & \textbf{40.00} & \textbf{52.40} & \textbf{78.67} & \textbf{59.80}\\
    \midrule
        & Chatbot & 16.67 & 74.70 & 35.60 & 49.20 & 73.33 & 49.90\\
        & Reasoner & 26.67 & 79.12 & 41.60 & 50.00 & 72.00 & 53.88\\
        \rowcolor{blue!8} \multirow{-3}{*}{\cellcolor{white}LLM-Debate~\cite{debate1}} & \textbf{X-MAS-Design} & \textbf{56.67} & \textbf{81.12} & \textbf{44.40} & \textbf{54.40} & \textbf{80.00} & \textbf{63.32}\\
    \midrule
        & Chatbot & 20.00 & 74.70 & 34.00 & 44.00 & 70.76 & 48.67\\
        & Reasoner & 40.00 & 76.31 & 42.40 & 45.60 & 68.00 & 54.46\\
        \rowcolor{blue!8} \multirow{-3}{*}{\cellcolor{white}DyLAN~\cite{dylan}} & \textbf{X-MAS-Design} & \textbf{63.33} & \textbf{80.32} & \textbf{42.80} & \textbf{46.80} & \textbf{76.00} & \textbf{61.85}\\
    \midrule
        & Chatbot & 23.33 & 72.69 & 34.80 & 44.80 & 68.00 & 48.72\\
        & Reasoner & 0.00 & 71.49 & 23.20 & 49.20 & 56.00 & 39.98\\
        \rowcolor{blue!8} \multirow{-3}{*}{\cellcolor{white}X-MAS-Proto} & \textbf{X-MAS-Design} & \textbf{70.00} & \textbf{79.12} & \textbf{47.20} & \textbf{52.80} & \textbf{76.00} & \textbf{65.02}\\
    \bottomrule
    \end{tabular}
    \vspace{-3mm}
    }
\end{table}

\textbf{Mixing chatbots and reasoners in X-MAS-Design achieves superior performance across domains and MAS methods.}
In Table~\ref{tab:x_mas_design_mixed}, we explore the potential of mixing chatbot and reasoner LLMs in X-MAS-Design.
From the table, we see that 
(1) X-MAS-Design, combining chatbot and reasoner agents powered by heterogeneous LLMs, consistently outperforms both standalone chatbot and reasoner configurations across all five domains.
(2) Standalone chatbot and reasoner configurations show complementary strengths and weaknesses.
The heterogeneous X-MAS-Design mitigates individual role limitations by combining chatbot and reasoner strengths, as guided by X-MAS-Bench’s 1.7 million evaluations.
This synergy enables robust performance across diverse domains.

\begin{wraptable}{R}{0.42\textwidth}
  \centering
  \setlength\tabcolsep{3pt}
  \vspace{-5mm}
  \caption{Examination on \textit{entirely new} benchmarks. X-MAS-Design achieves significantly best performance.}
    \label{tab:new_bench}
  \centering
  \begin{tabular}{ccc}
    \toprule
    Benchmark     &   AIME-25 & MATH-M \\
    \midrule
    Chatbot  &   13.33 &   14.18 \\
    Reasoner&   10.00 &   5.97 \\
    \rowcolor{blue!8} \textbf{X-MAS-Design}   &   \textbf{46.67} &   \textbf{48.13} \\
    \bottomrule
  \end{tabular}
    \vspace{-2mm}
\end{wraptable}
\textbf{Mixing chatbots and reasoners leads to dramatic improvements in math domain (AIME).}
We additionally evaluate homogeneous and heterogeneous MAS on two entirely new benchmarks: AIME-2025 and MATH-MAS in Table~\ref{tab:new_bench}.
From Table~\ref{tab:x_mas_design_mixed} and \ref{tab:new_bench}, we see that in math domain (i.e., AIME-2024, AIME-2025, MATH-MAS), X-MAS-Design contributes to substantial performance boosts.
Notably, for X-MAS-Proto, X-MAS-Design scores 70\% in AIME-2024, a 46.67\%-point gain over the second-best homogeneous MAS, indicating the potential of X-MAS in reasoning-intensive tasks.
Meanwhile, X-MAS-Design outperforms the second-best chatbot-based homogeneous MAS by 33\% and 34\% on the challenging AIME-2025 and MATH-MAS, respectively, indicating the generalization of our core idea.
In the era where reasoning models prevail, our experiments point out a potential direction: further scaling compute with X-MAS that mixes chatbots and reasoners.

\begin{figure}[t]
    \centering
    \includegraphics[width=1.0\linewidth]{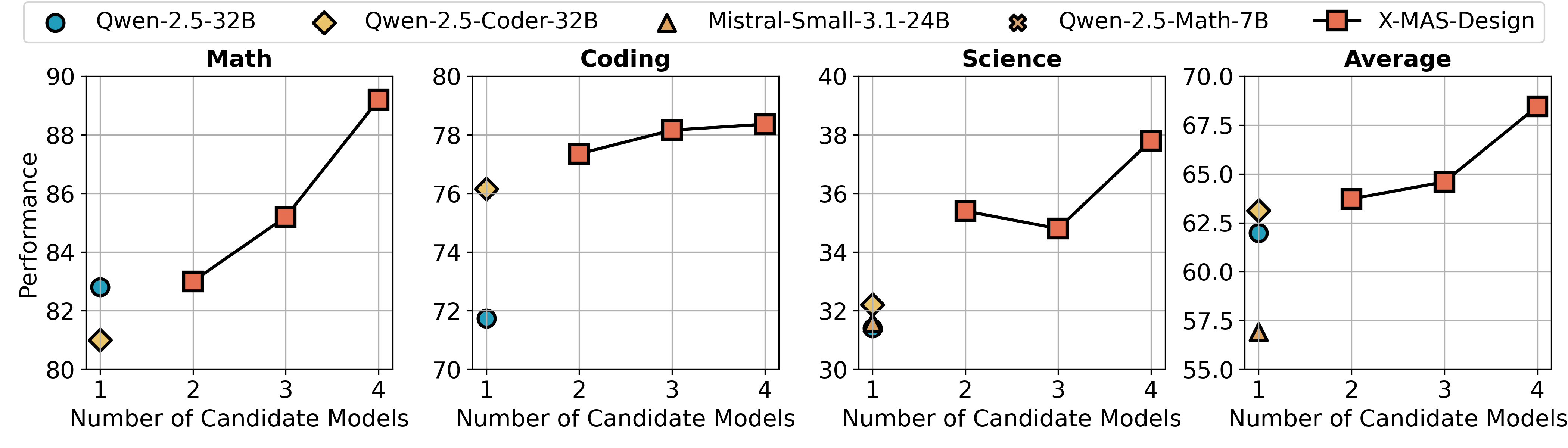}
    \caption{Diversity for the win. Experiments are conducted with X-MAS-Proto on three domains. Increasing the number of candidate models generally enhances the system performance, strongly indicating the benefits of LLM heterogeneity for MAS.}
    \vspace{-3mm}
    \label{fig:diversity_curve}
\end{figure}

\subsection{Ablation Study}

\textbf{Increasing the number of candidate models enhances the performance of X-MAS-Design.}
Following the setup in Section~\ref{sec:mas_experiments_chatbot}, we conduct experiments with X-MAS-Proto on three domains (math, coding, and science) by tuning the number of candidate models.
We use the full split for larger sample numbers.
From Figure~\ref{fig:diversity_curve}, we observe that
(1) X-MAS-Design consistently outperforms homogeneous MAS (i.e., 1 candidate model), indicating the benefits of X-MAS.
(2) With the number of candidate models increases, we can generally observe an increase of performance.
One exception is in the science domain, which can be attributed that the added model from 2 to 3 is not closely related to science.
This curve strongly indicates the benefits of including diverse LLMs in MAS.

\begin{wrapfigure}{r}{0.5\textwidth}
   \centering
   \vspace{-3mm}
    \includegraphics[width=0.5\textwidth]{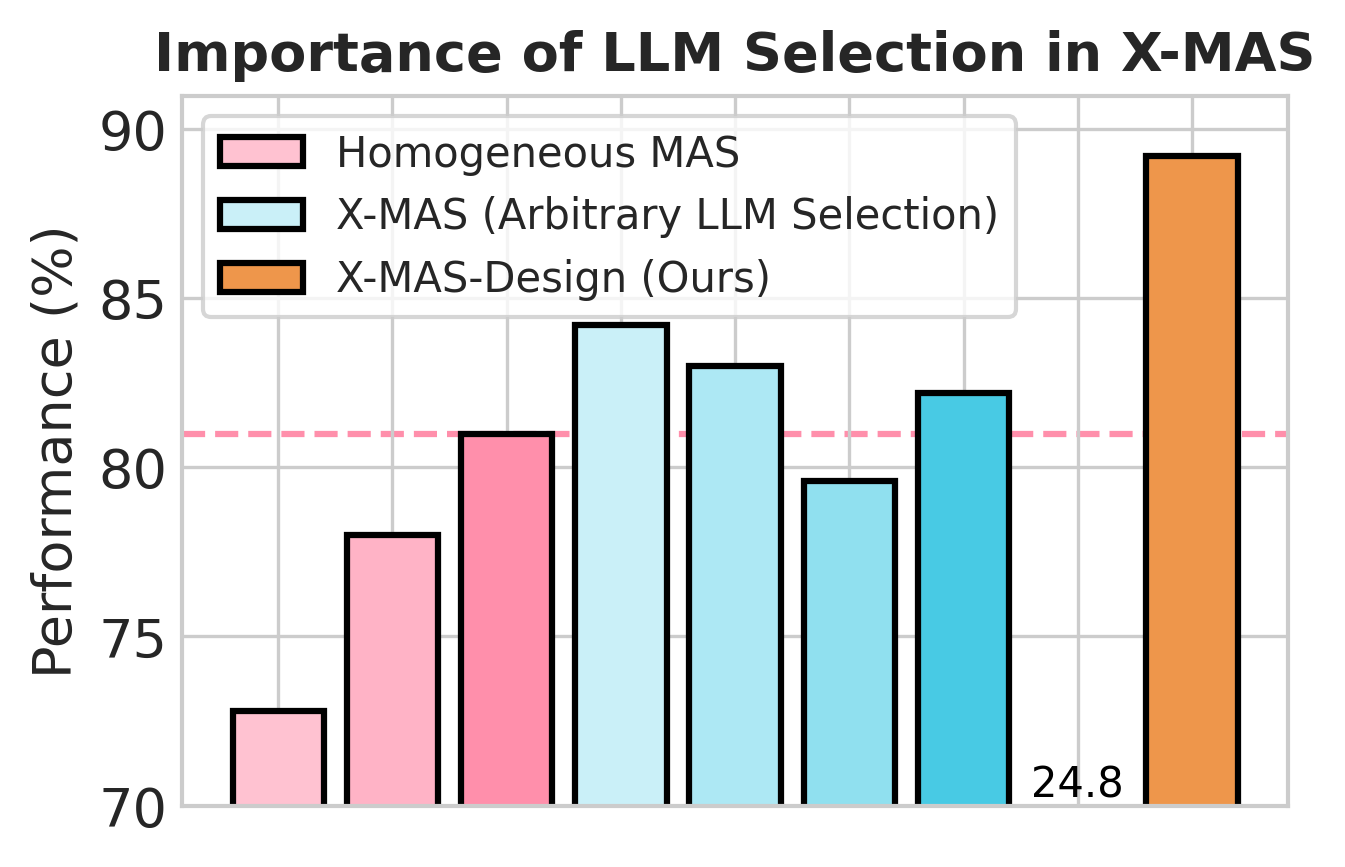}
    \vspace{-7mm}
    \caption{Comparing X-MAS with LLM selection guided by X-MAS-Bench and arbitrary selection. X-MAS-Design, which is guided by X-MAS-Bench, significantly performs the best.}
    \vspace{-3mm}
    \label{fig:x_mas_random_select}
\end{wrapfigure}
\textbf{Arbitrary model selection could lead to sub-optimal performance: X-MAS-Bench offers critical observations to guide the design of X-MAS.}
To verify the effectiveness of the observations from X-MAS-Bench, we compare X-MAS with LLM selection guided by X-MAS-Bench to X-MAS with arbitrary selection.
We follow the setup in Section~\ref{tab:x_mas_design_chatbot}, where we experiment on X-MAS-Proto on MATH-500.
We arbitrarily determine 5 reasonable sets of configurations for designing X-MAS (see details in Section~\ref{sec:x_mas_random_select}), denoted by \textcolor{blue!50}{blue bars} in Figure~\ref{fig:x_mas_random_select}.
Homogeneous MAS driven by three different LLMs is denoted by \textcolor{red!60}{red bars}.
From the figure, we see that
(1) X-MAS-Design, which is designed based on observations from X-MAS-Bench, significantly performs the best.
(2) Among those 5 X-MAS without X-MAS-Bench's guidance, 3 of them achieve slightly better performance than homogeneous MAS, while 1 performs slightly worse than the best homogeneous MAS and 1 even performs significantly worst (only 24.8\%).
This indicates that appropriate LLM selection is critical for ensuring the performance of X-MAS and that results in X-MAS-Bench can provide valuable insights.

\section{Conclusions}
\label{sec:conclusions}

This paper advocates building LLM-based MAS with heterogeneous LLMs.
We introduce X-MAS-Bench, a comprehensive testbed designed to assess the capabilities of various LLMs in supporting for MAS.
We provide a systematic empirical study, which assesses 27 LLMs (both chatbots and reasoners, both genralists and specialists) across 5 representative MAS-related functions and 5 common domains, highlighting the potential of employing heterogeneous LLMs in MAS.
Based on the insights from X-MAS-Bench, we examine the effects of transitioning from homogeneous to heterogeneous LLM-driven MAS (X-MAS-Design).
Our experiments operating on 4 MAS methods demonstrate that the performance of MAS can be significantly and consistently improved by leveraging heterogeneous MAS without any structural re-design, strongly supporting our advocacy.
See limitations in Section~\ref{sec:limitation}.

Our work highlights an intriguing direction that leverages the collective intelligence of heterogeneous LLMs to achieve higher-level intelligence without additional training.
Looking ahead, future research could explore areas such as automated or dynamic model selection, the impact of further scaling model candidates, optimizing the synergy between LLM selection and MAS, achieving strong performance with weaker agents, and training agents specifically suited for MAS.

\medskip
{
\small
\bibliographystyle{unsrt}
\bibliography{ref}
}


\appendix

\section{Limitations}
\label{sec:limitation}

Despite being the most comprehensive evaluation of LLMs for MAS, there are still LLMs that have not been included yet.
When transitioning from homogeneous to heterogeneous MAS, we currently rely on manual modification as a proof of concept.
Despite that the modification is quite simple, it is worthwhile to explore automated solutions.

\section{Broader Impacts}
\label{sec:broader_impacts}

This paper introduces X-MAS-Bench, aiming at assessing the capabilities of LLMs when being incorporated in MAS.
The assessed results and the corresponding findings could serve the community, facilitating appropriate model selections during the design of MAS.
Our X-MAS-Design aims to transition an existing homogeneous MAS to a heterogeneous one.
Similar concept could be extended to many existing MAS, making the overall system perform better.

The potential negative impacts of our approach mirror those associated with LLMs including risks of misuse.
However, these issues are intrinsic to LLM usage in general.

\section{Experimental Setups}
\label{app:experimental_setups}

We examine 27 LLMs, covering 20 chatbots (i.e., instructed LLMs) and 7 reasoners (i.e., reasoning LLMs).
Among the 20 chatbots, we consider general chatbots trained by different companies: Llama (Llama-3.1-8/70B-Instruct~\cite{dubey2024llama}), Qwen (Qwen2.5-7/14/32/72B-Instruct~\cite{qwen2.5}, Mistral (Mistral-7B-Instruct-v0.3~\cite{Mistral-7B-Instruct-v0.3}, Mistral-Small-3.1-24B-Instruct-2503~\cite{Mistral-Small-3.1-24B-Instruct-2503});
we also include domain-specific chatbots including mathematics (Qwen2.5-Math-7/72B-Instruct~\cite{qwen-math}), coding (Qwen2.5-Coder-7/14/32B-Instruct~\cite{Qwen2.5-coder}, Codestral-22B-v0.1~\cite{Codestral-22B-v0.1}), science (ChemDFM-v1.5-8B~\cite{chemdfm}, SciPhi-Mistral-7B-32k~\cite{SciPhi-Mistral-7B-32k}), medicine (Llama3-OpenBioLLM-70B~\cite{OpenBioLLMs} and BioMedGPT-LM-7B~\cite{luo2023biomedgpt}), and finance (Llama3-XuanYuan3-70B-Chat~\cite{Llama3-XuanYuan3-70B-Chat} and ZhiLu-2-8B-Instruct~\cite{ZhiLu-2-8B-Instruct})
The reasoners include LLMs from DeepSeek (DeepSeek-R1-Distill-Llama-8/70B and DeepSeek-R1-Distill-Qwen-7/14/32B~\cite{guo2025deepseek}), Qwen (QwQ-32B~\cite{qwq32b}) , other (OpenThinker-32B~\cite{openthoughts}) LLMs.
We set each model’s maximum token limit to its own capacity, 8192 tokens maxed, with a temperature of 0.5 by default.
Specially, all LLMs instantiated within the planning workflow are executed with their temperature fixed at 0 to guarantee as the planning involves format-following.
Our datasets cover domains including mathematics (AIME-2024~\cite{AIME-2024}, AQUA-RAT~\cite{aqua-rat}, GSM-Hard~\cite{gsm-hard}, MATH~\cite{hendrycksmath2021}, MMLU-Math~\cite{mmlu}, MMLU-Pro-Math~\cite{wang2024mmlupro}),
coding (HumanEval~\cite{humaneval}, HumanEval-Plus~\cite{evalplus}, MBPP~\cite{mbpp}, MBPP-Plus, MMLU-Coding, MMLU-Pro-coding),
science (GPQA-Main~\cite{rein2023gpqa}, GPQA-Diamond, SciBench~\cite{wangscibench}, SciEval~\cite{sun2024scieval}, SciKnowEval~\cite{feng2024sciknoweval}, MMLU-Sci, MMLU-Pro-Sci), 
medicine (MedMCQA~\cite{pal2022medmcqa}, MedQA~\cite{medqa}, PubMedQA~\cite{jin2019pubmedqa}, MMLU-Med, MMLU-Pro-Med),
and finance (FinanceBench~\cite{islam2023financebench}, FinQA~\cite{chen2021finqa}, FPB~\cite{fpb}, MMLU-Finan, MMLU-Pro-Finan), where each dataset is randomly sampled up to 500 examples without replacement (except for SciKnowEval, from which we draw 800 instances to ensure sufficient coverage of its specialized tasks).

\section{Experiments on X-MAS in Chatbot-Only Scenarios}
\label{app:experiments_mas_chatbot}

\subsection{Experiments setups of X-MAS-Design in Chatbot-Only Scenarios}

The available LLMs are Qwen-2.5-32B-Instruct, Qwen-2.5-Coder-32B-Instruct, Qwen-2.5-Math-7B-Instruct and Mistral-Small-3.1-24B-Instruct-2503.

\subsubsection{Agentverse}
\textbf{Mathematics.} The role assigner is Qwen-2.5-32B-Instruct, the solver is Qwen-2.5-Coder-32B-Instruct and Qwen-2.5-Math-7B-Instruct, the critic is Qwen-2.5-Coder-32B-Instruct, the evaluator is Qwen-2.5-32B-Instruct.

\textbf{Coding.} The role assigner is Qwen-2.5-32B-Instruct, the solver is Qwen-2.5-32B-Instruct, the critic is Qwen-2.5-Coder-32B-Instruct, the evaluator is Qwen-2.5-Coder-32B-Instruct.

\textbf{Science.} The role assigner is Qwen-2.5-32B-Instruct, the solver is Qwen-2.5-32B-Instruct, the critic is Qwen-2.5-32B-Instruct, the evaluator is Mistral-Small-3.1-24B-Instruct-2503.

\textbf{Medicine.} The role assigner is Qwen-2.5-32B-Instruct, the solver is Qwen-2.5-32B-Instruct, the critic is Mistral-Small-3.1-24B-Instruct-2503, the evaluator is Mistral-Small-3.1-24B-Instruct-2503.

\textbf{Finance.} The role assigner is Mistral-Small-3.1-24B-Instruct-2503, the solver is Qwen-2.5-Coder-32B-Instruct, the critic is Qwen-2.5-32B-Instruct, the evaluator is Mistral-Small-3.1-24B-Instruct-2503.

\subsubsection{LLM-Debate}
\textbf{Mathematics.} The debate agent is Qwen-2.5-Coder-32B-Instruct and Qwen-2.5-Math-7B-Instruct, the aggregator is Mistral-Small-3.1-24B-Instruct-2503.

\textbf{Coding.} The debate agent is Qwen-2.5-Coder-32B-Instruct, the aggregator is Mistral-Small-3.1-24B-Instruct-2503.

\textbf{Science.} The debate agent is Qwen-2.5-32B-Instruct, the aggregator is Qwen-2.5-32B-Instruct.

\textbf{Medicine.} The debate agent is Qwen-2.5-32B-Instruct, the aggregator is Mistral-Small-3.1-24B-Instruct-2503.

\textbf{Finance.} The debate agent is Qwen-2.5-Coder-32B-Instruct, the aggregator is Mistral-Small-3.1-24B-Instruct-2503.

\subsubsection{DyLAN}
\textbf{Mathematics.} The node agent is Qwen-2.5-Coder-32B-Instruct and Qwen-2.5-Math-7B-Instruct, the ranker is Mistral-Small-3.1-24B-Instruct-2503.

\textbf{Coding.} The node agent is Qwen-2.5-Coder-32B-Instruct, the ranker is Mistral-Small-3.1-24B-Instruct-2503.

\textbf{Science.} The node agent is Qwen-2.5-32B-Instruct, the ranker is Qwen-2.5-32B-Instruct.

\textbf{Medicine.} The node agent is Mistral-Small-3.1-24B-Instruct-2503, the ranker is Mistral-Small-3.1-24B-Instruct-2503.

\textbf{Finance.} The node agent is Qwen-2.5-Coder-32B-Instruct, the ranker is Mistral-Small-3.1-24B-Instruct-2503.

\subsubsection{X-MAS-Proto}
\textbf{Mathematics.} The planner is Qwen-2.5-32B-Instruct, the solver is Qwen-2.5-Coder-32B-Instruct and Qwen-2.5-Math-7B-Instruct, the reviser is Qwen-2.5-Coder-32B-Instruct, the evaluator is Qwen-2.5-32B-Instruct, the aggregator is Mistral-Small-3.1-24B-Instruct-2503.

\textbf{Coding.} The planner is Qwen-2.5-32B-Instruct, the solver is Qwen-2.5-Coder-32B-Instruct, the reviser is Qwen-2.5-Coder-32B-Instruct, the evaluator is Qwen-2.5-32B-Instruct, the aggregator is Mistral-Small-3.1-24B-Instruct-2503.

\textbf{Science.} The planner is Qwen-2.5-32B-Instruct, the solver is Qwen-2.5-32B-Instruct, the reviser is Qwen-2.5-32B-Instruct, the evaluator is Mistral-Small-3.1-24B-Instruct-2503, the aggregator is Qwen-2.5-32B-Instruct.

\textbf{Medicine.} The planner is Qwen-2.5-32B-Instruct, the solver is Qwen-2.5-32B-Instruct, the reviser is Mistral-Small-3.1-24B-Instruct-2503, the evaluator is Mistral-Small-3.1-24B-Instruct-2503, the aggregator is Mistral-Small-3.1-24B-Instruct-2503.

\textbf{Finance.} The planner is Mistral-Small-3.1-24B-Instruct-2503, the solver is Qwen-2.5-Coder-32B-Instruct, the reviser is Qwen-2.5-32B-Instruct, the evaluator is Mistral-Small-3.1-24B-Instruct-2503, the aggregator is Mistral-Small-3.1-24B-Instruct-2503.

\subsection{Experimental Setups of X-MAS with Non-X-MAS-Bench-Guided Model Selections}
\label{sec:x_mas_random_select}

We arbitrarily determine five reasonable manually designed model configurations to examine the robustness and performance sensitivity of the X-MAS-Design under diverse agent choices. These configurations are constructed without referring to the X-MAS-Bench, and are denoted as \texttt{X-MAS1} through \texttt{X-MAS5}. Each configuration includes distinct combinations of planner, solver, evaluator, reviser, and aggregator roles. For comparison, we also include the original X-MAS-Design configuration guided by X-MAS-Bench selection.

The X-MAS-Bench-guided configuration, referred to as \textbf{X-MAS-Design} in chatbot-only scenarios, adopts the following models for each agent role:
\begin{itemize}
    \item \textbf{Planner}: Qwen-2.5-32B-Instruct
    \item \textbf{Solver}: Qwen-2.5-Coder-32B-Instruct
    \item \textbf{Evaluator}: Qwen-2.5-32B-Instruct
    \item \textbf{Reviser}: Qwen-2.5-Coder-32B-Instruct
    \item \textbf{Aggregator}: Mistral-Small-3.1-24B-Instruct-2503
\end{itemize}
This configuration reflects a well-balanced assignment with domain-specialized solvers (e.g., math) and stronger general-purpose planning and evaluation agents.

In contrast, the five alternative configurations (\texttt{X-MAS1} to \texttt{X-MAS5}) are constructed based on general instruction-tuned LLMs without prior empirical optimization. These setups are:

\paragraph{X-MAS1}
\begin{itemize}
    \item \textbf{Planner}: Mistral-Small-3.1-24B-Instruct-2503
    \item \textbf{Solver}: Qwen-2.5-Math-7B-Instruct
    \item \textbf{Evaluator}: Qwen-2.5-Coder-32B-Instruct
    \item \textbf{Reviser}: Qwen-2.5-Math-7B-Instruct
    \item \textbf{Aggregator}: Qwen-2.5-32B-Instruct
\end{itemize}

\paragraph{X-MAS2}
\begin{itemize}
    \item \textbf{Planner}: Mistral-Small-3.1-24B-Instruct-2503
    \item \textbf{Solver}: Qwen-2.5-Coder-32B-Instruct
    \item \textbf{Evaluator}: Qwen-2.5-Math-7B-Instruct
    \item \textbf{Reviser}: Qwen-2.5-Coder-32B-Instruct
    \item \textbf{Aggregator}: Qwen-2.5-32B-Instruct
\end{itemize}

\paragraph{X-MAS3}
\begin{itemize}
    \item \textbf{Planner}: Qwen-2.5-Math-7B-Instruct
    \item \textbf{Solver}: Mistral-Small-3.1-24B-Instruct-2503
    \item \textbf{Evaluator}: Qwen-2.5-32B-Instruct
    \item \textbf{Reviser}: Mistral-Small-3.1-24B-Instruct-2503
    \item \textbf{Aggregator}: Qwen-2.5-Coder-32B-Instruct
\end{itemize}

\paragraph{X-MAS4}
\begin{itemize}
    \item \textbf{Planner}: Qwen-2.5-Coder-32B-Instruct
    \item \textbf{Solver}: Qwen-2.5-32B-Instruct
    \item \textbf{Evaluator}: Qwen-2.5-Math-7B-Instruct
    \item \textbf{Reviser}: Qwen-2.5-32B-Instruct
    \item \textbf{Aggregator}: Mistral-Small-3.1-24B-Instruct-2503
\end{itemize}

\paragraph{X-MAS5}
\begin{itemize}
    \item \textbf{Planner}: Qwen-2.5-32B-Instruct
    \item \textbf{Solver}: Qwen-2.5-Coder-32B-Instruct
    \item \textbf{Evaluator}: Mistral-Small-3.1-24B-Instruct-2503
    \item \textbf{Reviser}: Qwen-2.5-Coder-32B-Instruct
    \item \textbf{Aggregator}: Qwen-2.5-Math-7B-Instruct
\end{itemize}

All configurations are evaluated on the MATH-500 subset following the X-MAS-Proto scheme. The goal of this analysis is to understand the effect of heterogeneous agent assignments on final multi-agent performance, as well as to validate the necessity and advantages of X-MAS-Bench-guided agent selection. These baselines also serve to demonstrate the variance among manually configured pipelines in the absence of systematic design guidance.

\section{Experiments on X-MAS in Mixed Chatbot-Reasoner Scenarios}
\label{app:experiments_mas_mixed}

\subsection{Model Selections}

The available LLMs are Qwen-2.5-72B-Instruct and DeepSeek-R1-Distill-Qwen-32B.

\subsubsection{Agentverse}
\textbf{Mathematics.} The role assigner is Qwen-2.5-72B-Instruct, the solver is DeepSeek-R1-Distill-Qwen-32B, the critic is DeepSeek-R1-Distill-Qwen-32B, the evaluator is DeepSeek-R1-Distill-Qwen-32B.

\textbf{Coding.} The role assigner is Qwen-2.5-72B-Instruct, the solver is DeepSeek-R1-Distill-Qwen-32B, the critic is DeepSeek-R1-Distill-Qwen-32B, the evaluator is DeepSeek-R1-Distill-Qwen-32B.

\textbf{Science.} The role assigner is Qwen-2.5-72B-Instruct, the solver is DeepSeek-R1-Distill-Qwen-32B, the critic is DeepSeek-R1-Distill-Qwen-32B, the evaluator is DeepSeek-R1-Distill-Qwen-32B.

\textbf{Medicine.} The role assigner is Qwen-2.5-72B-Instruct, the solver is Qwen-2.5-72B-Instruct, the critic is Qwen-2.5-72B-Instruct, the evaluator is DeepSeek-R1-Distill-Qwen-32B.

\textbf{Finance.} The role assigner is Qwen-2.5-72B-Instruct, the solver is DeepSeek-R1-Distill-Qwen-32B, the critic is DeepSeek-R1-Distill-Qwen-32B, the evaluator is DeepSeek-R1-Distill-Qwen-32B.

\subsubsection{LLM-Debate}
\textbf{Mathematics.} The debate agent is DeepSeek-R1-Distill-Qwen-32B, the aggregator is DeepSeek-R1-Distill-Qwen-32B.

\textbf{Coding.} The debate agent is DeepSeek-R1-Distill-Qwen-32B, the aggregator is Qwen-2.5-72B-Instruct.

\textbf{Science.} The debate agent is DeepSeek-R1-Distill-Qwen-32B, the aggregator is DeepSeek-R1-Distill-Qwen-32B.

\textbf{Medicine.} The debate agent is Qwen-2.5-72B-Instruct, the aggregator is DeepSeek-R1-Distill-Qwen-32B.

\textbf{Finance.} The debate agent is DeepSeek-R1-Distill-Qwen-32B, the aggregator is DeepSeek-R1-Distill-Qwen-32B.

\subsubsection{DyLAN}
\textbf{Mathematics.}The node agent is DeepSeek-R1-Distill-Qwen-32B, the ranker is Qwen-2.5-72B-Instruct.

\textbf{Coding.} The node agent is DeepSeek-R1-Distill-Qwen-32B, the ranker is Qwen-2.5-72B-Instruct.

\textbf{Science.} The node agent is DeepSeek-R1-Distill-Qwen-32B, the ranker is Qwen-2.5-72B-Instruct.

\textbf{Medicine.} The node agent is Qwen-2.5-72B-Instruct, the ranker is Qwen-2.5-72B-Instruct.

\textbf{Finance.} The node agent is DeepSeek-R1-Distill-Qwen-32B, the ranker is Qwen-2.5-72B-Instruct.

\subsubsection{X-MAS-Proto}
\textbf{Mathematics.} The planner is Qwen-2.5-72B-Instruct, the solver is DeepSeek-R1-Distill-Qwen-32B, the reviser is DeepSeek-R1-Distill-Qwen-32B, the evaluator is DeepSeek-R1-Distill-Qwen-32B, the aggregator is DeepSeek-R1-Distill-Qwen-32B.

\textbf{Coding.} The planner is Qwen-2.5-72B-Instruct, the solver is DeepSeek-R1-Distill-Qwen-32B, the reviser is DeepSeek-R1-Distill-Qwen-32B, the evaluator is DeepSeek-R1-Distill-Qwen-32B, the aggregator is Qwen-2.5-72B-Instruct.

\textbf{Science.} The planner is Qwen-2.5-72B-Instruct, the solver is DeepSeek-R1-Distill-Qwen-32B, the reviser is DeepSeek-R1-Distill-Qwen-32B, the evaluator is DeepSeek-R1-Distill-Qwen-32B, the aggregator is DeepSeek-R1-Distill-Qwen-32B.

\textbf{Medicine.} The planner is Qwen-2.5-72B-Instruct, the solver is Qwen-2.5-72B-Instruct, the reviser is Qwen-2.5-72B-Instruct, the evaluator is DeepSeek-R1-Distill-Qwen-32B, the aggregator is DeepSeek-R1-Distill-Qwen-32B.

\textbf{Finance.} The planner is Qwen-2.5-72B-Instruct, the solver is DeepSeek-R1-Distill-Qwen-32B, the reviser is DeepSeek-R1-Distill-Qwen-32B, the evaluator is DeepSeek-R1-Distill-Qwen-32B, the aggregator is DeepSeek-R1-Distill-Qwen-32B.

\begin{figure}[t]
    \centering
    \includegraphics[width=1.0\linewidth]{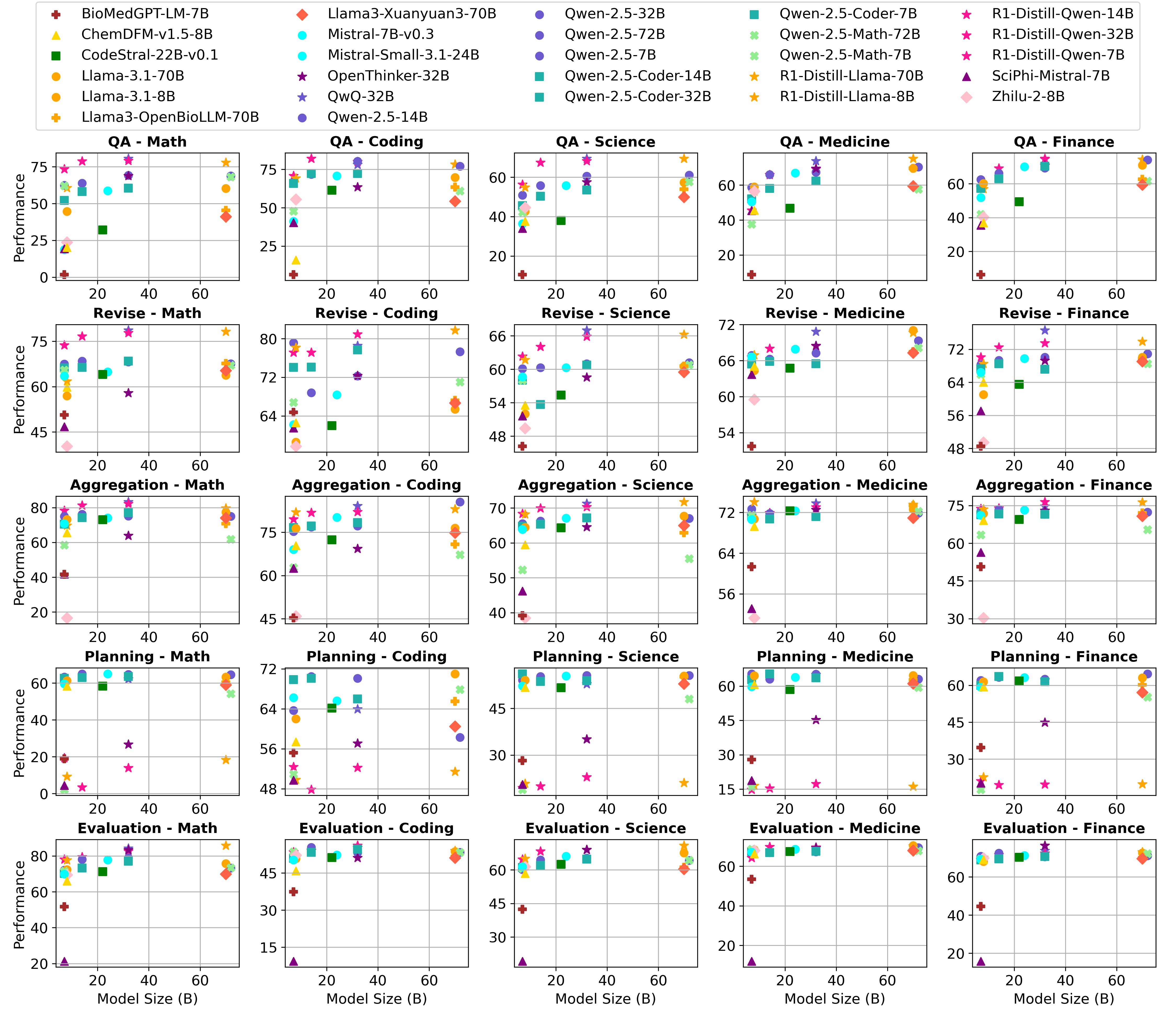}
    \caption{Benchmarking LLMs on 5 MAS-related functions and 5 domains.}
    \label{fig:bench_results_full}
\end{figure}

\begin{table}[htbp]
  \centering
  \caption{Top-3 Models per Function and Domain (reasoner and chatbot)}
  \label{tab:model-ranking_full}
  \renewcommand{\arraystretch}{1.2}
  \resizebox{\textwidth}{!}{
  \begin{tabular}{l c c c c c c}
    \toprule
    \textbf{Function} & \textbf{Rank}
     & \textbf{Mathematics} & \textbf{Coding} & \textbf{Science}
     & \textbf{Medicine}  & \textbf{Finance} \\
    \midrule

  \rowcolor{gray!10}
    & \goldmedal  & QwQ-32B (80.5)
                 & DeepSeek-R1-Distill-Qwen-14B (82.0)
                 & QwQ-32B (69.4)
                 & DeepSeek-R1-Distill-Llama-70B (75.1)
                 & DeepSeek-R1-Distill-Qwen-32B (74.8) \\
    \rowcolor{gray!10}  
    \textbf{QA}
    & \silvermedal& DeepSeek-R1-Distill-Qwen-32B (79.0)
                 & Qwen2.5-32B (80.3)
                 & DeepSeek-R1-Distill-Llama-70B (69.4)
                 & QwQ-32B (73.8)
                 & QwQ-32B (74.6) \\
    \rowcolor{gray!10}
    & \bronzemedal & DeepSeek-R1-Distill-Qwen-14B (78.8)
                  & DeepSeek-R1-Distill-Qwen-32B (80.0)
                  & DeepSeek-R1-Distill-Qwen-32B (68.3)
                  & Qwen2.5-72B (70.4)
                  & DeepSeek-R1-Distill-Llama-70B (74.3) \\
    \midrule

    & \goldmedal  & QwQ-32B (78.6)
                 & DeepSeek-R1-Distill-Llama-70B (81.7)
                 & QwQ-32B (67.0)
                 & Llama-3.1-70B (71.0)
                 & QwQ-32B (76.6) \\
    \textbf{Revise}
    & \silvermedal& DeepSeek-R1-Distill-Llama-70B (78.2)
                 & DeepSeek-R1-Distill-Qwen-32B (81.0)
                 & DeepSeek-R1-Distill-Llama-70B (66.3)
                 & DeepSeek-R1-Distill-Llama-70B (66.3)
                 & DeepSeek-R1-Distill-Llama-70B (73.9) \\
    & \bronzemedal & DeepSeek-R1-Distill-Qwen-32B (77.8)
                  & Qwen2.5-7B (79.2)
                  & DeepSeek-R1-Distill-Qwen-32B (65.9)
                  & DeepSeek-R1-Distill-Llama-70B (70.7)
                  & DeepSeek-R1-Distill-Qwen-32B (73.5) \\
    \midrule
    
    \rowcolor{gray!10}
    & \goldmedal  & QwQ-32B (83.2)
                 & Qwen2.5-72B (85.5)
                 & DeepSeek-R1-Distill-Llama-70B (71.7)
                 & DeepSeek-R1-Distill-Llama-8B (74.1)
                 & DeepSeek-R1-Distill-Qwen-32B (76.4) \\
    \rowcolor{gray!10}
    \textbf{Aggregation}
    & \silvermedal& DeepSeek-R1-Distill-Qwen-32B (82.2)
                 & QwQ-32B (84.2)
                 & QwQ-32B (71.3)
                 & QwQ-32B (73.8)
                 & DeepSeek-R1-Distill-Llama-70B (76.4) \\
    \rowcolor{gray!10}
    & \bronzemedal & DeepSeek-R1-Distill-Qwen-14B (81.2)
                  & DeepSeek-R1-Distill-Llama-70B (83.1)
                  & DeepSeek-R1-Distill-Qwen-32B (70.3)
                  & DeepSeek-R1-Distill-Llama-70B (73.6)
                  & QwQ-32B (74.6) \\
    \midrule

    & \goldmedal  & Qwen2.5-14B (65.0)
                 & Llama-3.1-70B (71.0)
                 & Qwen2.5-Coder-7B (56.1)
                 & Qwen2.5-Coder-14B (65.4)
                 & Qwen2.5-72B (64.7) \\
    \textbf{Planning}
    & \silvermedal& Mistral-Small-3.1-24B (65.0)
                 & Qwen2.5-14B (70.5)
                 & Qwen2.5-32B (55.6)
                 & Qwen2.5-7B (65.3)
                 & Qwen2.5-Coder-14B (63.6) \\
    & \bronzemedal & Qwen2.5-32B (64.7)
                  & Qwen2.5-32B (70.1)
                  & Qwen2.5-72B (55.6)
                  & Qwen2.5-32B (65.2)
                  & Qwen2.5-14B (63.2) \\
    \midrule

  \rowcolor{gray!10}
    & \goldmedal  & DeepSeek-R1-Distill-Llama-70B (85.9)
                 & DeepSeek-R1-Distill-Qwen-32B (56.2)
                 & DeepSeek-R1-Distill-Llama-70B (70.9)
                 & Llama-3.1-70B (70.5)
                 & OpenThinker-32B (76.6) \\
    \rowcolor{gray!10}
    \textbf{Evaluation}
    & \silvermedal& QwQ-32B (84.2)
                 & Qwen2.5-14B (55.4)
                 & DeepSeek-R1-Distill-Qwen-32B (69.1)
                 & DeepSeek-R1-Distill-Llama-70B (70.2)
                 & QwQ-32B (73.8) \\
    \rowcolor{gray!10}
    & \bronzemedal & OpenThinker-32B (83.3)
                  & QwQ-32B (55.3)
                  & OpenThinker-32B (69.0)
                  & DeepSeek-R1-Distill-Qwen-14B (69.8)
                  & DeepSeek-R1-Distill-Llama-70B (73.1) \\
    \bottomrule
  \end{tabular}
  }
\end{table}

\end{document}